%% file: AnonymousSubmission2027.tex
\documentclass[letterpaper]{article} 
\usepackage[draft]{aaai2027} 
\usepackage[hyphens]{url} 
\usepackage{graphicx} 
\usepackage{natbib} 
\usepackage{caption} 
\usepackage{amsmath}
\usepackage{booktabs}
\usepackage{xcolor}
\usepackage{tcolorbox}

\newif\ifdraftwithappendix
\draftwithappendixtrue 

\title{Thinking Under Uncertainty: Evidence Use and Information-Seeking in Language Models}
\author{Hua-Dong Xiong$^{1}$\thanks{Equal contribution; listed alphabetically},
Xinyuan Yan$^{2}$\footnotemark[1],
Li Ji-An$^{3}$,
Jingming Xue$^{1}$,
Marcelo Mattar$^{3}$,
Robert Wilson$^{1,4}$\thanks{Senior author}
}
\affiliations{
$^{1}$ School of Psychological and Brain Sciences, Georgia Tech \\
$^{2}$ Department of Neurosurgery, Baylor College of Medicine \\
$^{3}$ Department of Psychology, New York University \\
$^{4}$ Center of Excellence for Computational Cognition, Georgia Tech
}

\begin{document}

\maketitle

\begin{abstract}
Inference-time thinking improves the performance of large language models, but aggregate outcomes do not reveal whether models use available evidence more effectively or seek information that could improve future decisions. We distinguish these responses by measuring action preference, thinking length, and reported confidence under matched uncertainty. Ten open-weight models completed matched horizon-style two-armed bandit trials in thinking and non-thinking modes. A cognitive model separated value-guided action and uncertainty-independent choice noise from two behavioral signatures of exploration: a UCB-like preference for the less-known arm and Thompson-like choice variability that increases with total uncertainty. On average, thinking strengthened value-guided action and reduced uncertainty-independent choice noise, without producing UCB-like exploration or strengthening Thompson-like exploration. Outside action, the information-imbalanced history condition, which also displayed more observations than the matched balanced condition, was associated with greater thinking length. Reported confidence became more sensitive to decision difficulty and more strongly associated with chosen task evidence. We interpret these thinking-length and reported-confidence patterns as consistent with metacognitive control and metacognitive monitoring, respectively, without establishing either process. Decoder sweeps, especially temperature, altered choice noise and thinking length but did not reproduce the joint cross-output pattern. In this controlled decision setting, thinking improved how models acted on current evidence, while neither measured signature supported a shift toward a more information-seeking policy.
\end{abstract}

\section{Introduction}

Inference-time thinking has emerged as a second scaling axis for language models, complementing the scaling of model parameters and training data \citep{brown_language_2020}. From chain-of-thought prompting to reasoning models, generating intermediate reasoning before an answer often improves performance \citep{wei_chainofthought_2022,openai_openai_2024,wang_selfconsistency_2023}. Yet aggregate performance does not reveal whether thinking helps models use available evidence more effectively or seek information that could improve future decisions. These responses need not move together: a model may act more consistently on what it already knows without acting to learn more. We therefore distinguish them by asking how thinking changes action preference, thinking length, and reported confidence under matched uncertainty.

These outputs play different functional roles under uncertainty. At the action level, the distinction between evidence use and information seeking parallels the exploration--exploitation problem: whether a model follows the option best supported by current evidence or chooses an action that may improve its information state. Choice variability alone, however, does not establish information seeking: a model can appear exploratory because its choices are noisy, even when uncertainty has no structured association with action. We therefore separate value-guided choice, two exploration signatures, and an uncertainty-independent noise floor. A UCB-like signature appears when a model prefers the less-known arm, whereas a Thompson-like signature appears when choice variability increases with total uncertainty. We subsequently refer to these components as UCB-like and Thompson-like exploration. These behavioral analogies do not imply that a model internally implements either algorithm. Information-seeking exploration therefore differs both from output stochasticity and from search within a thinking trace \citep{guiomar_reasoning_2026}.

Thinking length and reported confidence capture other responses: a model can generate a longer thinking trace or change its reported confidence without choosing an action that reduces uncertainty. Uncertainty-responsive thinking length is consistent with metacognitive control, and evidence-sensitive reported confidence is consistent with metacognitive monitoring, but neither pattern establishes the corresponding process \citep{hay_selecting_2014,lee_trading_2021}. These nonexclusive accounts therefore make different cross-output predictions. Behavioral narrowing predicts more consistent value-guided action; an information-seeking policy predicts UCB-like exploration or stronger Thompson-like exploration; metacognitive control and monitoring predict uncertainty-responsive changes in thinking length and reported confidence, respectively.

We use a controlled decision task to separate these functional components. Psychology-derived tasks characterize language-model behavioral profiles \citep{binz_using_2023,coda-forno_cogbench_2024}, while fitted cognitive models reveal interpretable value trade-offs and hypothesis updating \citep{murthy_cognitive_2025,xiong_hypothesis_2026}. Building on this approach, ten models from the Gemma 4 \citep{gemma_2026}, GPT-OSS \citep{openai_gptoss120b_2025}, Nemotron 3 \citep{nvidia_nvidia_2025}, and Qwen 3.5/3.6 \citep{qwen_qwen35_2026} families completed matched horizon-style two-armed bandit trials \citep{wilson_humans_2014} in thinking and non-thinking modes. We measured how the same value and uncertainty structure shaped action logits, thinking length, and reported confidence. The cognitive decomposition estimates UCB-like and Thompson-like exploration alongside value-guided action and uncertainty-independent choice noise \citep{gershman_deconstructing_2018}.

We test these predictions in sequence. We first ask whether thinking narrows action around the available evidence or supports a more information-seeking policy through UCB-like exploration or stronger Thompson-like exploration \citep{zhao_echo_2025}. We then ask how the same task variables predict thinking length and reported confidence. Finally, decoder sweeps test a simpler sampling account: whether changing temperature, top-\(p\), or top-\(k\) reproduces the same cross-output pattern.

Thinking strengthened value-guided action, while neither measured signature supported a shift toward a more information-seeking policy. Outside action, information-imbalanced histories were associated with greater thinking length, while reported confidence became more sensitive to decision difficulty and task evidence.

\section{Methods}

\subsection{Task and Prompting Procedure}

\begin{figure}[htbp]
\centering
\includegraphics[width=\columnwidth]{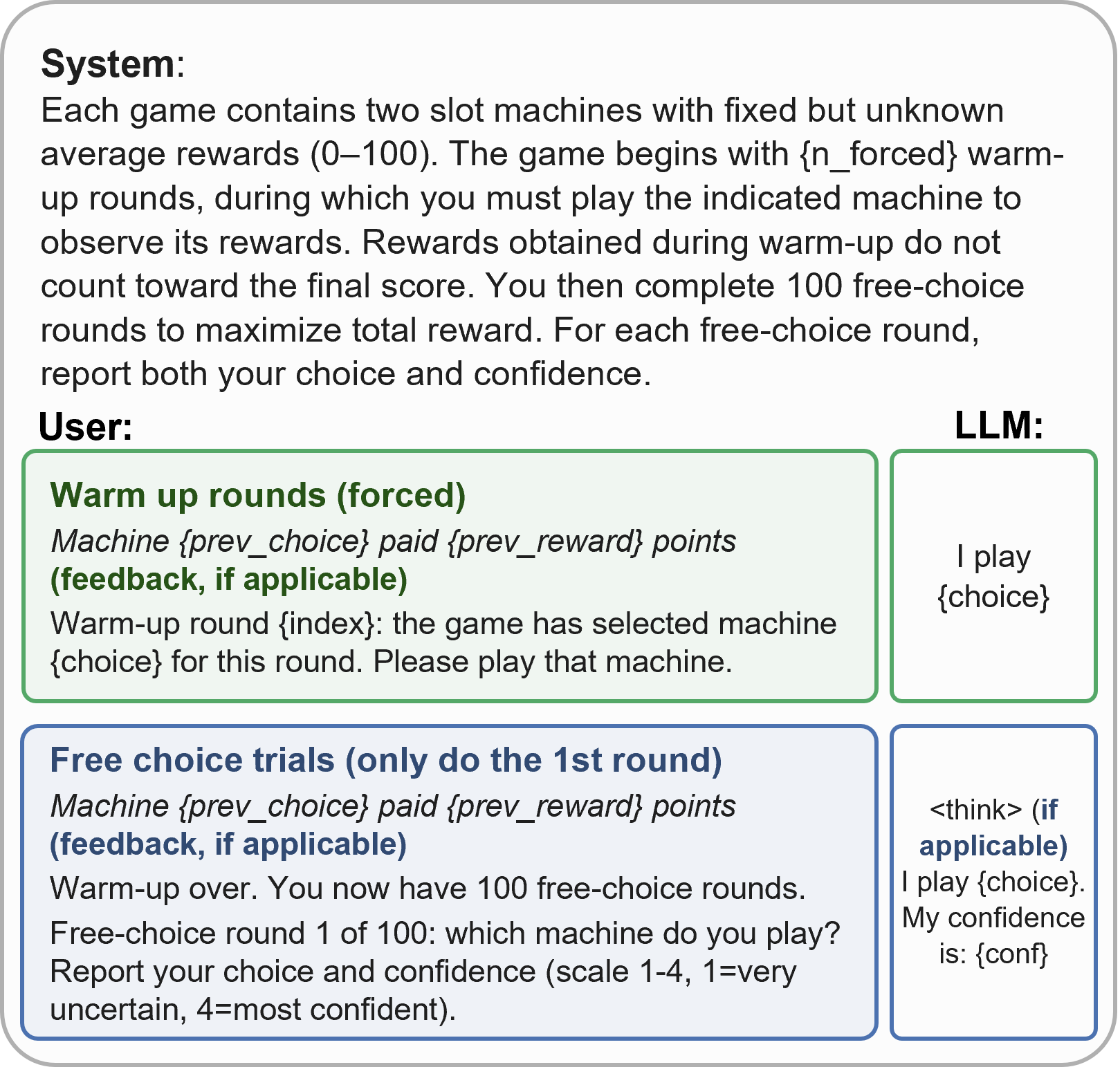}
\caption{Horizon bandit task. The model observed controlled warm-up rewards from two machines, then chose a machine and reported confidence. Thinking mode allowed a private thinking trace before this response. We analyzed the initial free choice.}
\label{fig:task}
\end{figure}

We adapted the horizon-based two-armed bandit paradigm \citep{wilson_humans_2014} into a multi-turn dialogue (Fig.~\ref{fig:task}). In each game, the model was told that two slot machines had fixed but unknown mean rewards on a 0--100 scale. It first observed a controlled warm-up history from each machine, then chose a machine and reported confidence on a four-point scale. Each game was framed as a 100-round decision horizon, giving the initial choice potential informational value for later decisions. We executed and analyzed only that initial choice, before self-selected outcomes could alter the evidence state. This controlled decision provides the cleanest test of whether the model follows current evidence or acts to improve its information state.

We constructed candidate warm-up histories by sampling rewards from a Gaussian distribution centered at 50 with standard deviation 10, rounding each reward to an integer, and clipping it to the 0--100 range. We used three observation-count conditions, \((1,1)\), \((3,3)\), and \((2,6)\), where each pair gives the warm-up counts for the two arms. Comparing \((1,1)\) with \((3,3)\) changed total uncertainty while keeping the arms equally observed, whereas comparing \((3,3)\) with \((2,6)\) changed relative uncertainty while matching total uncertainty. For each condition, we divided the absolute difference between the two empirical arm means into five one-point bins from zero to five. Within each bin, we set 40 evenly spaced target differences and selected the candidate history closest to each target. In the information-imbalanced condition, we counterbalanced both the side of the less-observed arm and whether it had the higher empirical mean.

Each model completed the same 600 trials in thinking and non-thinking modes. Thinking mode allowed the model to generate a private thinking trace before answering, whereas non-thinking mode answered directly. We recorded action logits over the two machine choices, confidence logits over the four report levels, and thinking length, defined as the number of generated thinking tokens. We summarized the confidence logits as reported confidence on a common four-point scale. The Technical Supplement provides further experimental details and measurement definitions.

\subsection{Computational Modeling}

\begin{figure}[htbp]
\centering
\includegraphics[width=\columnwidth]{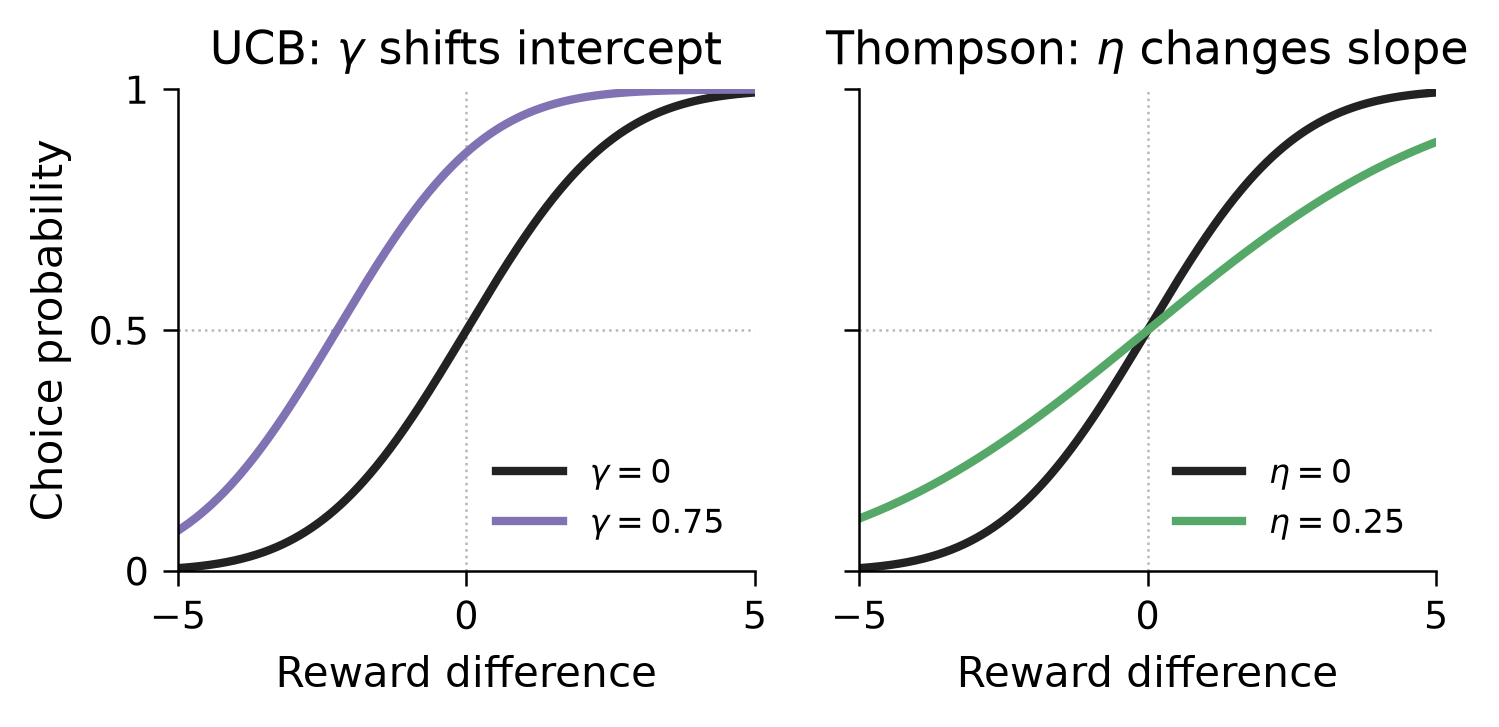}
\caption{Schematic behavioral signatures of UCB-like and Thompson-like exploration under the choice-level model. When the left arm is less observed (\(RU>0\)), increasing \(\gamma\) translates the choice curve toward that arm (left). Increasing \(\eta\) makes choices more variable as total uncertainty grows and flattens the choice curve around indifference (right). The uncertainty-independent noise parameter \(\lambda\) is held fixed in both panels; the curves are illustrative rather than fitted data.}
\label{fig:model-mechanisms}
\end{figure}

For arm \(a\), let \(\bar r_a\) denote the empirical mean reward and \(n_a\) the number of warm-up observations. We defined the standard-error proxy \(s_a=10/\sqrt{n_a}\), value difference \(V=\bar r_L-\bar r_R\), relative uncertainty \(RU=s_L-s_R\), and total uncertainty \(TU=\sqrt{s_L^2+s_R^2}\). Let \(a_i\in\{L,R\}\) denote the action on trial \(i\). For interpretation, we write the decomposition at the choice level as
\begin{equation}
P(a_i=L)=\Phi\!\left(
\frac{V_i+\gamma RU_i}
{\sqrt{\lambda^2+\eta^2TU_i^2}}
\right).
\label{eq:choice-model}
\end{equation}
Here \(\Phi\) is the standard normal cumulative distribution function. A positive \(\gamma\) shifts choice toward the less-observed arm, producing a UCB-like signature. The parameter \(\eta\) makes choice variability increase with total uncertainty, producing a Thompson-like signature. These components constitute UCB-like and Thompson-like exploration, respectively, whereas \(\lambda\) sets an uncertainty-independent noise floor \citep{gershman_deconstructing_2018}. The analogies describe behavioral effects rather than internal algorithms or optimal policies. Fig.~\ref{fig:model-mechanisms} isolates the effects of \(\gamma\) and \(\eta\) on the choice curve.

We fitted this model to the action-logit difference between the two machines. Section 2.2 of the Technical Supplement provides the fitting procedure and model diagnostics.

\section{Results}

\subsection{Thinking Strengthened Evidence Use without Evidence for a More Information-Seeking Policy}

\begin{figure}[htbp]
\centering
\includegraphics[width=\columnwidth]{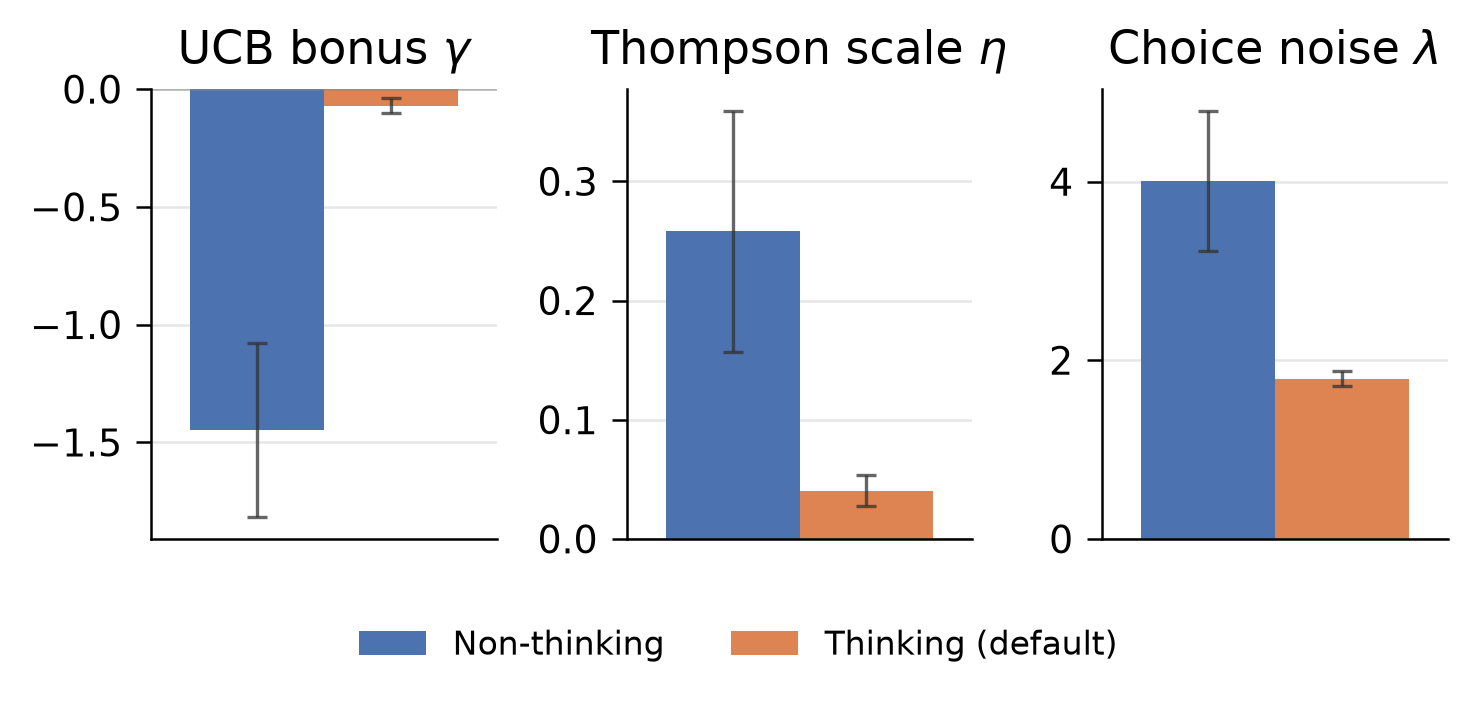}
\caption{Thinking-mode estimates show lower uncertainty-independent choice noise without UCB-like exploration or stronger Thompson-like exploration. Bars show mean model-level estimates of UCB-like exploration, Thompson-like exploration, and uncertainty-independent choice noise across the nine models that could be compared across modes. Error bars show model-level SEM. With thinking, models neither favor the less-known arm nor become more variable as total uncertainty increases, while uncertainty-independent choice noise is lower.}
\label{fig:cog_params}
\end{figure}

Thinking strengthened value-guided action without producing UCB-like exploration or strengthening Thompson-like exploration. It shifted the UCB-like coefficient from a strong negative estimate toward zero rather than into a preference for the less-known arm, while the Thompson-like coefficient decreased, indicating less, not more, variability as total uncertainty grew. Defining \(\Delta\theta=\theta_{\text{thinking}}-\theta_{\text{non-thinking}}\) across the nine comparable models, the mean changes were \(\Delta\gamma=1.38\), \(\Delta\eta=-0.22\), and \(\Delta\lambda=-2.21\) in value-equivalent units (Fig.~\ref{fig:cog_params}). The positive \(\Delta\gamma\) moved mean \(\gamma\) from \(-1.45\) to \(-0.07\), not above zero; mean \(\eta\) decreased from \(0.26\) to \(0.04\), and mean \(\lambda\) from \(4.01\) to \(1.79\).

Value dominated action selection in thinking mode, whereas relative and total uncertainty contributed little to the fitted action preferences. Non-thinking actions tracked value less clearly and varied more across models. One model could not be compared across modes because its non-thinking choices did not track value, and the remaining model-level estimates were heterogeneous. Because uncertainty-independent variability can make aggregate behavior look exploratory without reflecting a systematic response to uncertainty, this heterogeneity does not identify an exploration signature. The cross-mode evidence therefore shows lower uncertainty-independent choice noise and clearer value-guided action under thinking, while neither measured exploration signature supports a shift toward a more information-seeking policy. The Technical Supplement reports the individual-model exploration estimates. We next test how the same task variables predict thinking length and reported confidence.

\subsection{Thinking Length Tracked Information-Imbalanced Histories, While Reported Confidence Tracked Decision Difficulty and Task Evidence}

\begin{figure*}[t]
\centering
\includegraphics[width=0.94\textwidth]{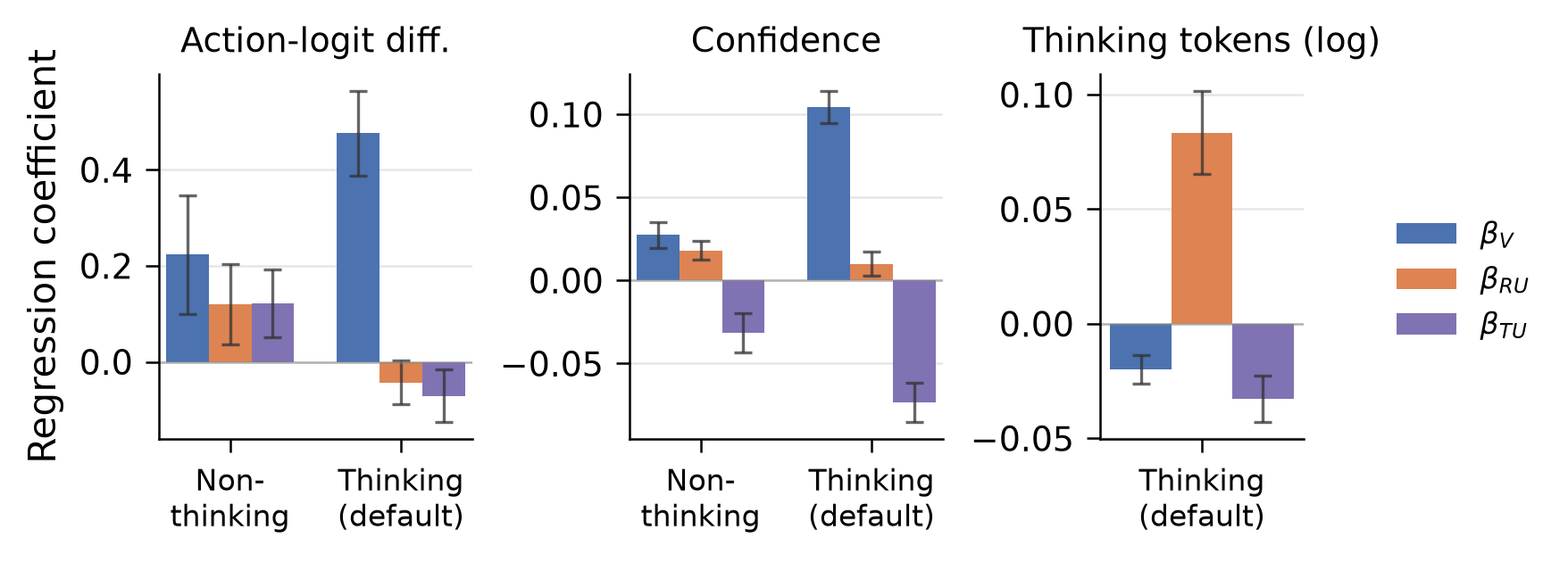}
\caption{Task variables are associated more clearly with thinking length and reported confidence, whereas value dominates choice. The three outcomes are the magnitude of the action-logit difference (left), reported confidence (middle), and log-transformed thinking length (right). Bar colors identify the coefficients for the magnitude of value difference, the magnitude of relative uncertainty, and centered total uncertainty, as shown in the legend. Bars show means across ten models. Error bars show model-level SEM. The thinking-length regression includes only thinking mode because non-thinking produces no thinking trace.}
\label{fig:behavior-regressions}
\end{figure*}

The same task variables could predict thinking length and reported confidence even when the measured action signatures did not support a more information-seeking policy. We therefore regressed the magnitude of the action-logit difference, thinking length, and reported confidence on the magnitude of value difference (\(\lvert V\rvert\)), the magnitude of relative uncertainty (\(\lvert RU\rvert\)), and total uncertainty. In this design, \(\lvert RU\rvert\) distinguishes the information-imbalanced \((2,6)\) histories from the balanced histories, whereas total uncertainty captures the overall uncertainty of the decision (Fig.~\ref{fig:behavior-regressions}).

\paragraph{Thinking length.}
Thinking length was most clearly associated with information-imbalanced histories, represented by the magnitude of relative uncertainty, \(\lvert RU\rvert\). In this design, however, that regressor marks the \((2,6)\) histories, which contained two more displayed observations than the balanced \((3,3)\) histories with matched total uncertainty. It therefore identifies an association with the information-imbalanced history condition rather than an isolated effect of information imbalance. Thinking length was greater after these histories, whereas the magnitude of value difference and total uncertainty had little association with length; this increase did not correspond to a bias toward the less-known arm. This task-responsive association is consistent with metacognitive control, but it does not establish that models explicitly monitored uncertainty or actively allocated additional thinking.

\begin{figure}[htbp]
\centering
\includegraphics[width=0.9\columnwidth]{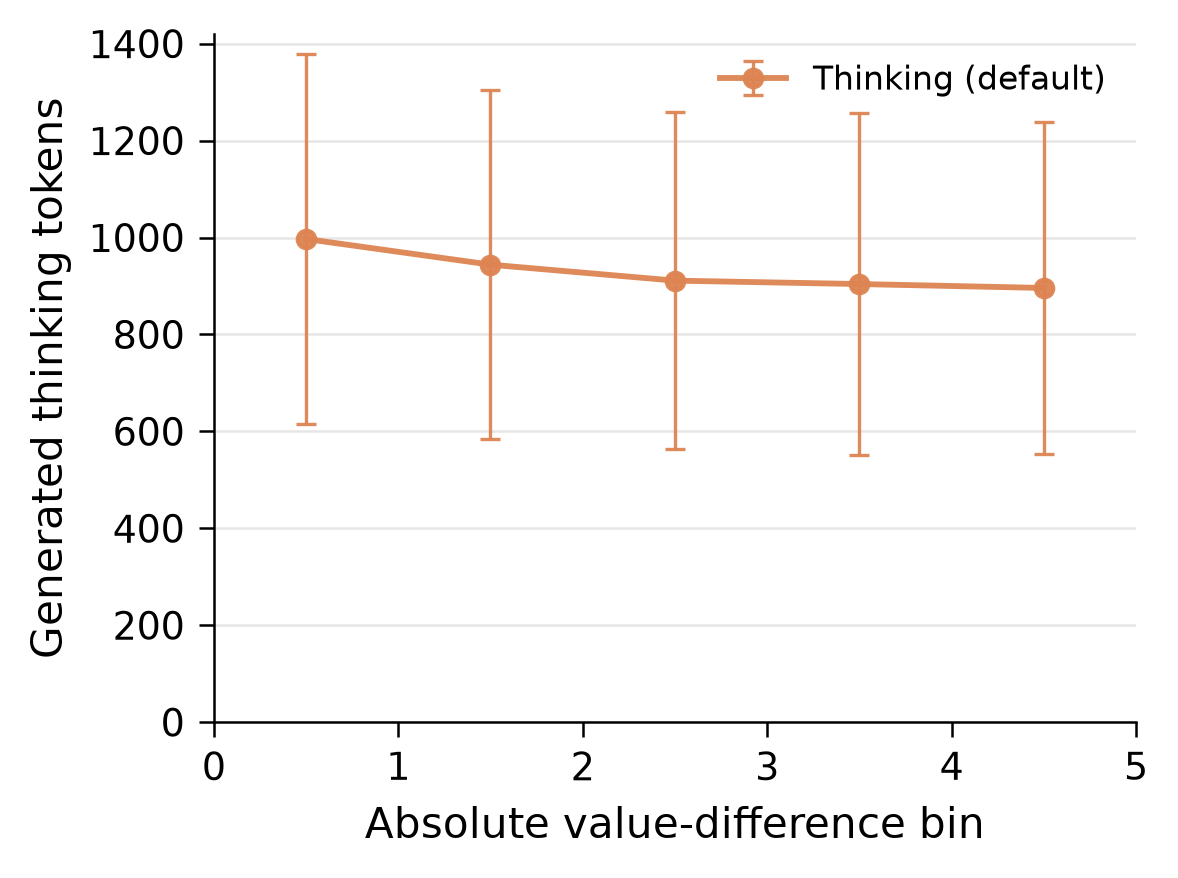}
\caption{Thinking length increases only weakly with decision difficulty. Points show mean thinking length across ten models by bins of the magnitude of value difference. Thinking length is slightly greater for the hardest decisions, but the bins overlap substantially. Error bars show model-level SEM.}
\label{fig:thinking-length}
\end{figure}

Decision difficulty showed a weaker pattern. Harder value comparisons produced only a small descriptive increase in thinking length, with wide variation across models and substantial overlap across bins (Fig.~\ref{fig:thinking-length}); we therefore treat this trend as suggestive. The pattern is not well summarized by a generic ``harder decisions take longer'' account: although the comparison remains descriptive, information-imbalanced histories showed a clearer association with thinking length than value similarity did.

\begin{figure}[htbp]
\centering
\includegraphics[width=0.9\columnwidth]{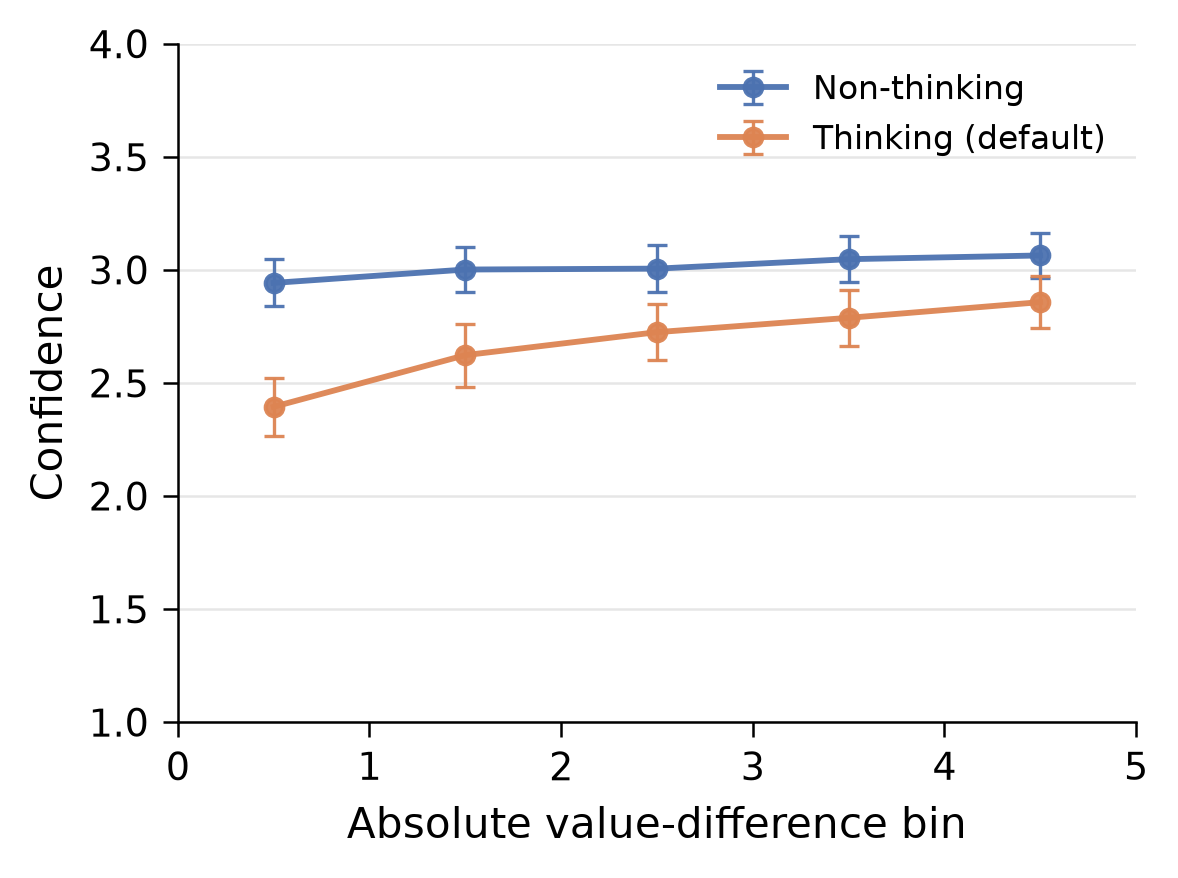}
\caption{Thinking increases reported-confidence sensitivity to decision difficulty. Mean reported confidence across ten models is shown across bins of the magnitude of value difference, from hardest to easiest decisions. Error bars show model-level SEM.}
\label{fig:confidence-difficulty}
\end{figure}

\paragraph{Reported confidence.}
Thinking made reported confidence more sensitive to the magnitude of value difference and total uncertainty. These variables captured complementary features of the task: the magnitude of value difference indexed discriminability between the arms, whereas total uncertainty indexed how uncertain the evidence remained across both arms. Reported confidence rose more strongly with the magnitude of value difference and declined more strongly with total uncertainty, becoming higher when the arms were easier to distinguish and lower when the decision was more uncertain overall. Only thinking mode showed a strong descriptive gradient across decision difficulty (Fig.~\ref{fig:confidence-difficulty}). Non-thinking reported confidence was nearly flat across the five bins, whereas thinking-mode reported confidence was lower overall and rose as decisions became easier. Thinking therefore did not merely lower reported confidence; it selectively reduced it when the evidence was ambiguous and increased it as one option became clearly better.

\begin{figure}[htbp]
\centering
\includegraphics[width=\columnwidth]{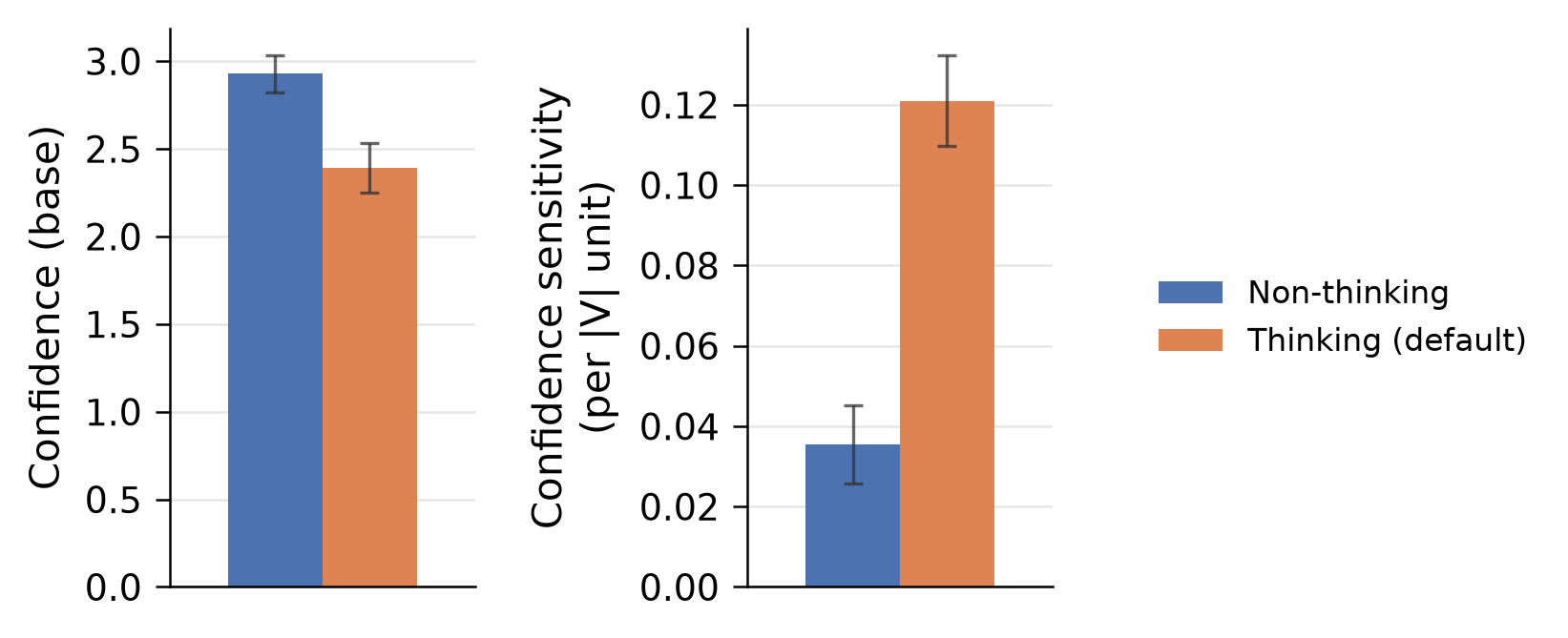}
\caption{Thinking lowers baseline reported confidence and increases its sensitivity to decision difficulty. Bars show mean model-level baseline reported confidence and sensitivity to the magnitude of value difference. Error bars show model-level SEM.}
\label{fig:confidence-difficulty-fit}
\end{figure}

Model-level fits separated two components of this pattern (Fig.~\ref{fig:confidence-difficulty-fit}): thinking lowered baseline reported confidence and increased its sensitivity to the magnitude of value difference. A lower baseline alone could reflect a general conservative response; the steeper slope shows that the mode difference was difficulty-sensitive rather than uniform. The modes therefore differed most on hard decisions and partially converged as the magnitude of value difference grew. This directional shift held for all ten checkpoints, although its size varied across models. The Technical Supplement reports the individual-model fits.

The reported-confidence pattern could reflect sensitivity to external task evidence, the model's own measured decision state, or both. We therefore characterized the two marginal associations separately by correlating reported confidence with chosen task evidence and with the action-logit margin between the selected and unselected actions.

\begin{figure}[htbp]
\centering
\includegraphics[width=0.8\columnwidth]{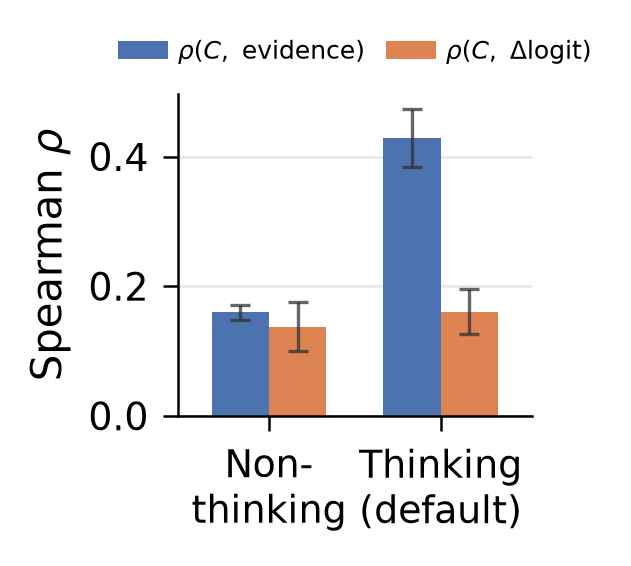}
\caption{Reported confidence was more strongly associated with chosen task evidence in thinking mode, while its separately estimated association with the action-logit margin changed little. Bars show mean within-model Spearman correlations between reported confidence and chosen evidence or the chosen action-logit margin across ten models. Error bars show model-level SEM.}
\label{fig:metacognition}
\end{figure}

Analyzed separately, the mean within-model Spearman correlation between reported confidence and chosen task evidence increased from \(0.16\) without thinking to \(0.43\) with thinking, whereas the correlation with the chosen action-logit margin changed from \(0.14\) to \(0.16\) (Fig.~\ref{fig:metacognition}). The stronger evidence association is consistent with metacognitive monitoring, but these marginal correlations do not establish that the mode difference was specific to task evidence, calibration to correctness, or direct access to a latent decision state.

Under thinking, value dominated action; information-imbalanced histories were associated with greater thinking length; and the magnitude of value difference and total uncertainty more strongly predicted reported confidence. Neither measured action signature supported a more information-seeking policy. This cross-output pattern could still arise from how the thinking trace was sampled rather than from a broader functional effect of thinking, which the decoder sweeps test next.

\subsection{Decoder Stochasticity Did Not Reproduce the Joint Pattern}

\begin{figure}[htbp]
\centering
\includegraphics[width=\columnwidth]{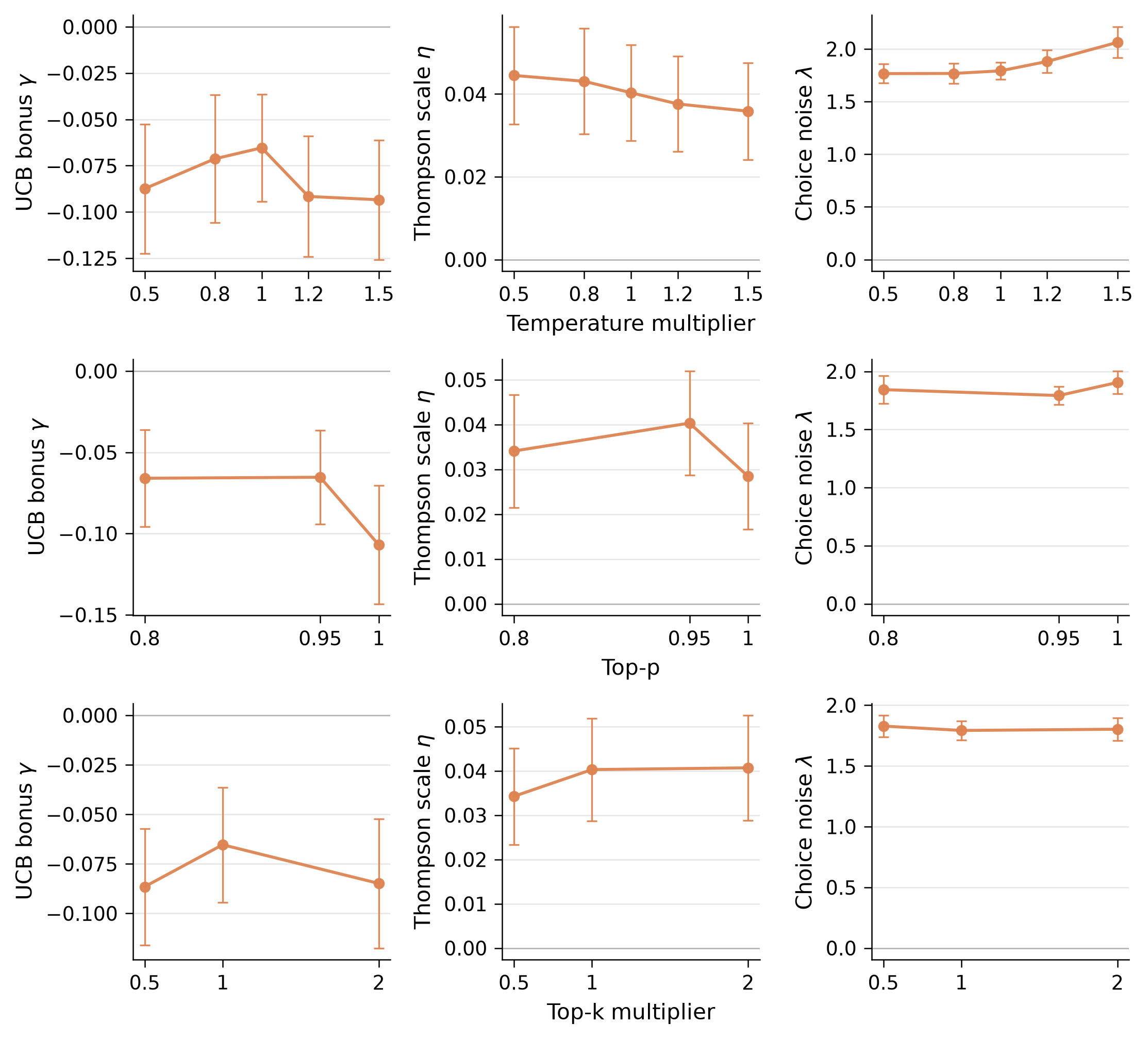}
\caption{Decoding settings move choice noise more than structured exploration. Columns show UCB-like exploration (a preference for the less-known arm), Thompson-like exploration (choice variability that increases with total uncertainty), and uncertainty-independent choice noise; rows show temperature, top-\(p\), and top-\(k\) varied only during private thinking generation. Means are computed across ten models. Error bars show model-level SEM. The exploration effects remain limited, and the clearest change is greater choice noise at higher thinking-sampling temperatures.}
\label{fig:decoding-cognitive}
\end{figure}

\begin{figure}[htbp]
\centering
\includegraphics[width=\columnwidth]{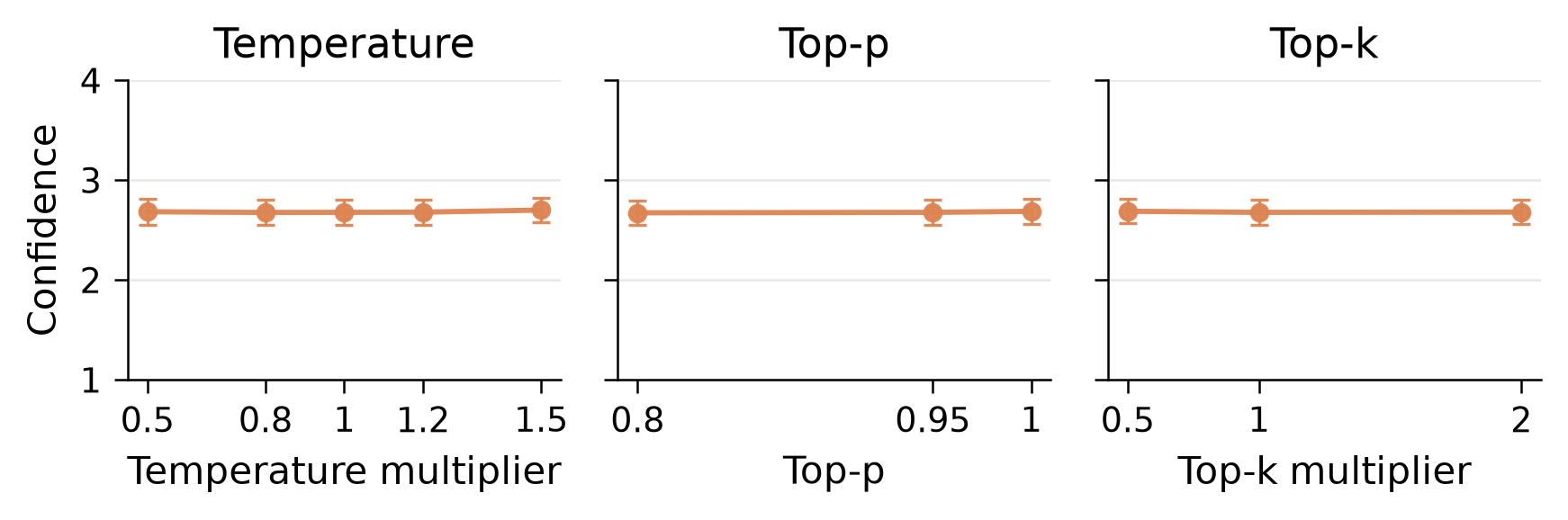}
\caption{Reported confidence is largely insensitive to decoding settings. Mean reported confidence is shown when temperature, top-\(p\), and top-\(k\) were varied only during private thinking generation. Means are computed across ten models. Error bars show model-level SEM.}
\label{fig:decoding-confidence}
\end{figure}

\begin{figure}[htbp]
\centering
\includegraphics[width=\columnwidth]{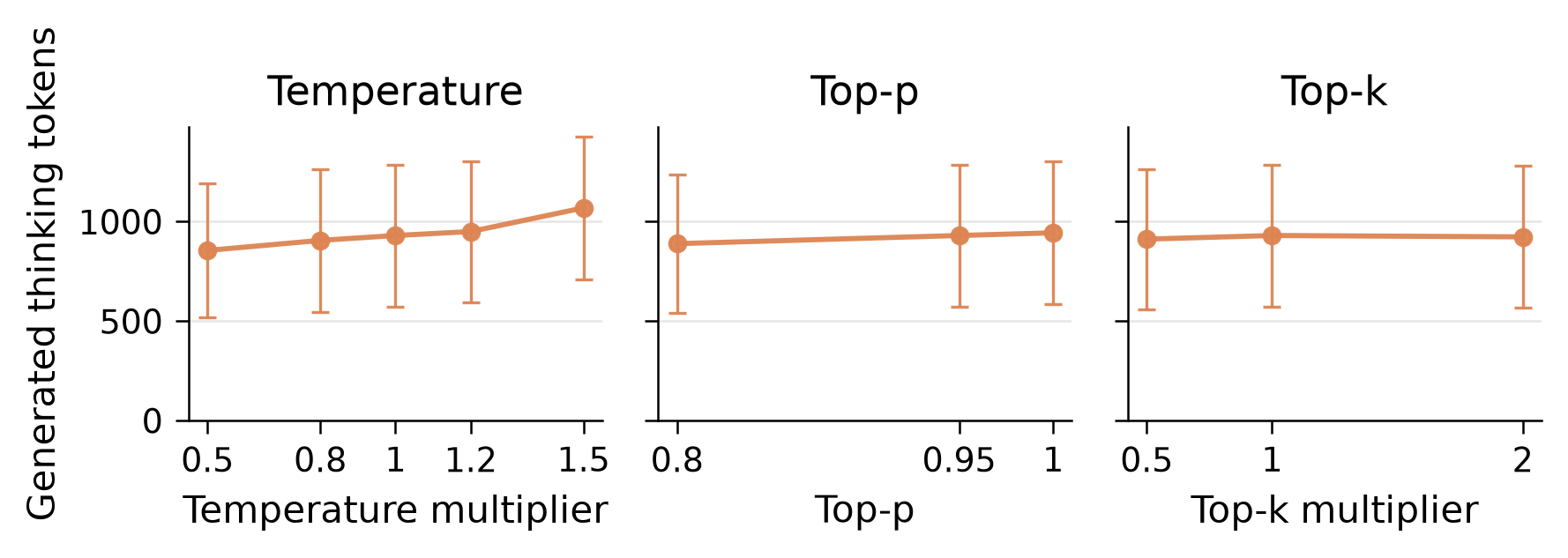}
\caption{Thinking length responds more to temperature and top-\(p\) than to top-\(k\). Mean thinking length is shown when each setting was varied only during private thinking generation. Thinking length increases across temperature, rises more modestly across top-\(p\), and remains approximately flat across top-\(k\). Means are computed across ten models. Error bars show model-level SEM.}
\label{fig:decoding-thinking-length}
\end{figure}

Decoder settings can alter output variability without necessarily changing how uncertainty guides action. If the preceding cross-output pattern arose simply from how the thinking trace was sampled, however, changing the decoder during thinking generation should reproduce it. We therefore varied temperature, top-\(p\), and top-\(k\) one at a time while generating the private thinking trace, then re-estimated the exploration components. Higher temperature increased uncertainty-independent downstream choice noise and thinking length, with the length increase present in all ten models. Top-\(p\) produced a smaller rise in thinking length without a consistent noise trend, whereas top-\(k\) had little effect. Across all three sweeps, reported confidence changed little, and initial choices did not become more likely to favor the less-known arm or more variable as total uncertainty increased (Figs.~\ref{fig:decoding-cognitive}--\ref{fig:decoding-thinking-length}).

Changes in the sampled thinking trace can therefore propagate into uncertainty-independent downstream choice variability. This downstream noise does not directly measure decoder entropy. Because the action logits were read without the sweep transformations, the effect arose through the sampled thinking trace rather than direct rescaling or truncation of the action distribution. Even at the highest tested temperature, choice noise remained below the non-thinking level. Within the tested ranges, the decoder sweeps did not reproduce the broader non-thinking choice pattern or the joint changes across outputs. Temperature nevertheless lengthened the thinking trace without corresponding changes in reported confidence or the uncertainty-specific action components. Thinking length alone was therefore insufficient to account for the cross-mode pattern.

\section{Discussion}

\paragraph{Better use of evidence is not better exploration.}
Thinking made choices more strongly tied to observed value and reduced uncertainty-independent choice noise, while neither measured signature supported a more information-seeking policy. The models showed neither reliable UCB-like nor substantial Thompson-like exploration. Lower uncertainty-independent choice noise can support more reliable use of current evidence, but reduced randomness is not itself evidence of information seeking. Because the non-thinking estimates were heterogeneous and one model's non-thinking choices did not track value, we cannot conclude that thinking removed a previously identifiable exploration strategy. The narrower conclusion is that thinking produced clearer value-guided actions without adding a positive information-seeking signature. This conclusion is compatible with behavioral narrowing \citep{zhao_echo_2025} at the measured decision, but it leaves open whether thinking adds other strategies or changes behavior across longer trajectories.

\paragraph{Thinking length, reported confidence, and information-seeking action are distinct outcomes.}
Task variables were associated with thinking length and reported confidence even though the measured action signatures did not support a more information-seeking policy: information-imbalanced histories were associated with thinking length, while reported confidence tracked decision difficulty and task evidence. An action-only analysis would miss these associations, whereas a trace-only analysis could mistake greater thinking length for information seeking. Greater thinking length and evidence-acquiring action are therefore distinct behavioral outcomes. These associations were output-specific rather than globally increased or suppressed. This length pattern is consistent with metacognitive control, and the stronger association between reported confidence and chosen evidence is consistent with metacognitive monitoring, but neither pattern establishes the corresponding process. Prior work asks whether language models can predict their own correctness \citep{kadavath_language_2022}, express calibrated uncertainty \citep{xiong_can_2024,yoon_reasoning_2025}, or monitor and control internal activations \citep{ji-an_language_2025}. Related work indicates that verbal confidence can contain answer-quality information beyond token log probabilities \citep{kumaran_how_2026}. Here, thinking length measures the generated trace, and we did not estimate the benefit or cost needed to establish rational control of thinking \citep{shenhav_expected_2013,hay_selecting_2014,lieder_resourcerational_2020}. Reported-confidence sensitivity likewise does not establish calibration to correctness or direct access to a latent decision state.

The decoder sweeps narrow a simpler explanation. Sampling changes propagated through the thinking trace and altered downstream noise and length, but the tested settings did not reproduce the joint pattern across action, thinking length, and reported confidence. Trace length and output variability are therefore not sufficient summaries of the functional response to uncertainty. These sweeps do not exclude every sampling-based account.

\paragraph{Implications for epistemic action.}
Some uncertainty can be reduced by further processing the current context, whereas other uncertainty requires new evidence. In retrieval, question asking, experimentation, scientific discovery, and planning, better use of available evidence cannot substitute for actions that change the information state. Our controlled decision does not establish behavior across richer trajectories; it isolates the distinction between using available evidence and acting to acquire more evidence.

\paragraph{Limitations.}
Richer and changing environments provide essential evidence about how LLMs learn over time, but they make it difficult to identify which component of behavior changes because self-selected outcomes continually alter the evidence state. We therefore use a deliberately simple, stationary two-option task and analyze the initial choice after a controlled warm-up history. Holding the evidence state fixed allows us to separate value-guided action, UCB-like and Thompson-like exploration, uncertainty-independent choice noise, thinking length, and reported confidence under matched stimuli. This control comes with a clear scope condition: although the announced \(H=100\) created a long-horizon incentive for information gathering, we did not execute later choices, so the results characterize the initial decision rather than learning across an extended sequence of self-selected actions. Human participants commonly show a preference for the less-known option in comparable horizon-bandit tasks \citep{wilson_humans_2014,gershman_deconstructing_2018}, but we did not test humans with the same dialogue implementation, prompts, reward sequences, and confidence measure. A matched human experiment would be required to establish human correspondence.

The information-imbalanced history also displayed more rewards than the balanced \((3,3)\) history with matched total uncertainty, so its association with thinking length cannot isolate information imbalance from the amount of evidence presented. Thinking length and reported confidence are behavioral observables rather than direct readouts of hidden processes: thinking length measures the generated trace rather than all internal computation, and reported confidence may partly reflect learned reporting conventions rather than latent certainty \citep{ji-an_language_2025,miao_closing_2026}. Finally, aggregate uncertainty in the figures reflects variation across models, not posterior uncertainty within a single hierarchical population model. These boundaries constrain generalization to extended trajectories and direct mechanistic interpretation, while the controlled design still supports a behavioral separation among action, thinking length, and reported confidence.

\paragraph{Conclusion.}
Thinking and information-seeking action are distinct responses to uncertainty. In our controlled decision setting, thinking improved the use of available evidence, while neither measured signature supported a more information-seeking policy. Evaluations of reasoning agents should therefore measure evidence acquisition separately from answer quality, output consistency, and thinking length.

\clearpage
\bibliography{references}

\ifdraftwithappendix
\clearpage
\appendix
\input{TechnicalSupplementBody}
\fi

\end{document}

%% file: TechnicalSupplementBody.tex
\section{Experimental Details}

\subsection{Models and Thinking Modes}

We evaluated ten open-weight models from four families: GPT-OSS \citep{openai_gptoss120b_2025}, Gemma 4 \citep{gemma_2026}, Nemotron 3 \citep{nvidia_nvidia_2025}, and Qwen 3.5/3.6 \citep{qwen_qwen35_2026}. Table~\ref{tab:models} lists the evaluated variants.

Each model completed the same trials in thinking and non-thinking modes. Thinking mode generated a thinking trace before the answer, whereas non-thinking mode answered directly. The prompts, trial histories, legal actions, and action measurements were matched across modes.

\begin{table}[t]
\centering
\footnotesize
\begin{tabular}{@{}ll@{}}
\toprule
Model family & Evaluated variants \\
\midrule
GPT-OSS & 20B, 120B \\
Gemma 4 & E2B, E4B, A4B \\
Nemotron 3 & Nano A3B, Super A12B \\
Qwen & 3.5 4B, 3.5 9B, 3.6 A3B \\
\bottomrule
\end{tabular}
\caption{Model families and variants included in the reported analyses.}
\label{tab:models}
\end{table}

\subsection{Horizon Task Construction}
\label{sec:construction}

Following the horizon task \citep{wilson_humans_2014}, every trial was framed as a fixed decision horizon of \(H=100\) free-choice rounds. This long horizon gave the initial choice potential informational value for many later decisions, approximating a continuing bandit without asking the model to interpret an abstract infinite horizon. Collection and analysis stopped after that initial decision, before self-selected outcomes could change the evidence state. This controlled choice provides the cleanest test of whether the model follows current evidence or acts to improve its information state.

The three warm-up observation-count conditions, \((1,1)\), \((3,3)\), and \((2,6)\), formed a minimal L-shaped design. Comparing \((1,1)\) with \((3,3)\) changed total uncertainty while holding relative uncertainty at zero; comparing \((3,3)\) with \((2,6)\) changed relative uncertainty while matching total uncertainty. These comparisons provided separate contrasts for the two uncertainty variables. However, because the information-imbalanced \((2,6)\) histories displayed eight rewards whereas the balanced \((3,3)\) histories displayed six, the relative-uncertainty contrast also changed history length; its association with thinking length cannot isolate information imbalance from the amount of displayed evidence. Each condition contained 200 trials distributed evenly across five empirical absolute value-difference bins, \([0,1)\), \([1,2)\), \([2,3)\), \([3,4)\), and \([4,5]\), yielding 40 trials per bin and 600 default trials per model and mode. Rewards were rounded and clipped draws centered at 50 with standard deviation 10 and support from 0 to 100. Histories were assigned to bins using the empirical difference between the displayed arm means.

The information-imbalance condition was counterbalanced across the left and right arms. Trial construction also balanced the sign of value difference, the left--right position of the machine letters, letter mention order, reward order, and recency. Consequently, relative uncertainty was not systematically confounded with value, screen position, a particular letter, or the most recent reward. Every mode and generation condition used the same trial set.

\subsection{Task Prompts}
\label{sec:prompts}

Let \(N\in\{2,6,8\}\) denote the total number of warm-up observations and let \(S\) denote one of the two counterbalanced confidence mappings below. The following system-message schema states the fixed horizon directly as 100 rounds.

\begin{tcolorbox}[colback=gray!5!white, colframe=black!75!white, title=Horizon Task System Message, fonttitle=\bfseries, fontupper=\footnotesize\ttfamily]
\label{lst:system-prompt}
\textbf{<System>}\\
You are playing slot-machine games in a casino. Each game has two slot\\
machines, each labeled with a capital letter, and every play pays a whole\\
number of points between 0 and 100. Each machine has a fixed average payoff,\\
but individual payoffs vary randomly.\\
Each game starts with \(N\) warm-up rounds. In each warm-up round, the\\
game selects which machine you play, and you respond exactly:\\
I play <LETTER>\\
These rounds let you observe both machines, but their points do not count\\
toward your score.\\
After the warm-up rounds, you have 100 free-choice rounds.\\
Your goal is to earn as many points as possible across these rounds.\\
During free-choice rounds, you select the machine yourself and also report\\
your confidence that your choice will help you earn more points in this\\
game. Confidence uses a 4-point scale, where\\
\(S\). Respond in exactly this form:\\
I play <LETTER>.\\
My confidence is: <NUMBER>
\end{tcolorbox}

For \(i=1,\ldots,N\), let \(x_i\in\{A,B\}\) denote the machine assigned on warm-up round \(i\) and let \(r_i\) denote its reward. The dialogue accumulated a fixed observation history through the following turn structure.

\begin{tcolorbox}[colback=gray!5!white, colframe=black!75!white, title=Forced-History and Free-Choice Turn Templates, fonttitle=\bfseries, fontupper=\footnotesize\ttfamily]
\label{lst:turn-prompts}
\textbf{First forced turn}\\
\textbf{<User>}\\
Warm-up round 1: the game has selected machine \(x_1\) for this\\
round. Please play that machine.\\
\textbf{<Assistant>}\\
I play \(x_1\)

\vspace{0.8em}
\textbf{Subsequent forced turn}\\
\textbf{<User>}\\
Machine \(x_{i-1}\) paid \(r_{i-1}\) points.\\
Warm-up round \(i\): the game has selected machine \(x_i\) for this\\
round. Please play that machine.\\
\textbf{<Assistant>}\\
I play \(x_i\)

\vspace{0.8em}
\textbf{First free-choice turn}\\
\textbf{<User>}\\
Machine \(x_N\) paid \(r_N\) points.\\
The warm-up rounds are over. You now have 100 free-choice rounds in this\\
game. Free-choice round 1 of 100: which machine do you play, \(A\) or \(B\)?\\
Report your choice and confidence (scale 1-4) using the\\
required format.
\end{tcolorbox}

The confidence-scale direction alternated within each ordered task cell. One of the following sentences replaced \(S\) in the system message.

\begin{tcolorbox}[colback=gray!5!white, colframe=black!75!white, title=Counterbalanced Confidence Instructions, fonttitle=\bfseries, fontupper=\footnotesize\ttfamily]
\label{lst:confidence-prompt}
\textbf{One-is-high instruction}\\
1 = most confident, 2 = somewhat confident, 3 = somewhat uncertain, and 4 = very uncertain

\vspace{0.8em}
\textbf{Four-is-high instruction}\\
1 = very uncertain, 2 = somewhat uncertain, 3 = somewhat confident, and 4 = most confident
\end{tcolorbox}

At the initial free choice, we measured the probability assigned to each legal machine letter and each confidence value rather than parsing a free-form response.

\subsection{Thinking-Generation Manipulations}

The default thinking condition used temperature \(1.00\), top-\(p=0.95\), and a maximum thinking length of 4096 tokens. Its top-\(k\) was 50 for GPT-OSS and Nemotron 3, 64 for Gemma 4, and 20 for Qwen. Non-thinking trials generated no thinking trace.

Within thinking mode, we varied one generation parameter at a time. Temperatures were \(\{0.50,0.80,1.00,1.20,1.50\}\), and top-\(p\) values were \(\{0.80,0.95,1.00\}\). Top-\(k\) values were \(\{25,50,100\}\) for GPT-OSS and Nemotron 3, \(\{32,64,128\}\) for Gemma 4, and \(\{10,20,40\}\) for Qwen. These manipulations applied only to the generated thinking trace. At the answer boundary, we recorded the raw conditional log probabilities of the two legal actions and their signed action-logit difference. The thinking-generation parameters were not applied to these action scores. We sampled the recorded continuation action from a separate fixed-\(T=1\) conditional softmax over the two scores, without top-\(k\) or top-\(p\) truncation.

\section{Measurement and Analysis Details}

\subsection{Action Preference, Reported Confidence, and Thinking Length}

Let \(\log p_i(L)\) and \(\log p_i(R)\) denote the raw conditional log probabilities assigned to the two legal actions at the answer boundary. The primary action measure was their signed action-logit difference,
\[
d_i=\log p_i(L)-\log p_i(R),
\]
which is also the difference between the corresponding logits because their common normalization term cancels.
The cognitive-model analysis used \(d_i\), and the behavioral regression used its magnitude \(\lvert d_i\rvert\). Thus, the reported action results retained graded preferences rather than reducing each trial to a generated response.

For reported confidence, let \(q_i(c)\) be the normalized probability assigned to response \(c\in\{1,2,3,4\}\). After reversing the one-is-high condition, we expressed every trial on a common scale where four denotes greatest confidence and computed
\[
C_i=\sum_{c=1}^{4}c\,q_i(c).
\]
All reported confidence analyses use \(C_i\).

Thinking length was the number of generated thinking tokens. For regression, the outcome was \(\log(n_{\mathrm{thinking}}/100)\). Traces that reached the 4096-token limit were treated as right-censored observations rather than as completed traces.

Variation in thinking length across task conditions is consistent with metacognitive control, whereas associations between reported confidence and task variables are consistent with metacognitive monitoring. These behavioral observables do not establish hidden control or monitoring processes.

\subsection{Choice Model and Continuous-Logit Estimation}

\paragraph{Choice-level representation.}
Following cognitive-model analyses of language-model behavior \citep{binz_turning_2023,murthy_cognitive_2025}, we used a horizon-task decomposition of two exploration signatures \citep{gershman_deconstructing_2018}. For arm \(a\), let \(\bar r_a\) denote the empirical mean reward and \(n_a\) the number of warm-up observations. Let \(a_i\in\{L,R\}\) denote the action on trial \(i\). We defined
\begin{equation}
\begin{aligned}
s_a&=\frac{10}{\sqrt{n_a}},&
V&=\bar r_L-\bar r_R,\\
RU&=s_L-s_R,&
TU&=\sqrt{s_L^2+s_R^2}.
\end{aligned}
\label{eq:task-variables}
\end{equation}
The choice-level hybrid model was
\begin{equation}
P(a_i=L)=\Phi\!\left(
\frac{V_i+\gamma RU_i}
{\sqrt{\lambda^2+\eta^2TU_i^2}}
\right),
\label{eq:choice-probit}
\end{equation}
where \(\Phi\) is the standard normal cumulative distribution function. The numerator combines reward value with a relative-uncertainty shift: when \(RU>0\), a positive \(\gamma\) favors the less-observed left arm, producing a UCB-like signature. The denominator combines uncertainty-independent choice noise \(\lambda\) with variability that increases with total uncertainty through \(\eta TU\), producing a Thompson-like signature. These components constitute UCB-like and Thompson-like exploration, respectively. Thus, \(\gamma\) shifts the choice curve, whereas \(\eta\) flattens it more strongly when total uncertainty is high. These are behavioral analogies, not claims that the model internally implements either algorithm.

\paragraph{Reported continuous-logit estimator.}
The reported parameter estimates use the graded preference retained in \(d_i\) rather than reducing each trial to one sampled choice. The continuous observation model was
\begin{equation}
d_i \sim \mathcal{N}\!\left(
\beta_V V_i+\beta_{RU}RU_i,\,
\lambda_d^2+\eta_d^2TU_i^2
\right),
\label{eq:hybrid}
\end{equation}
with normalized parameters
\begin{equation}
\gamma=\frac{\beta_{RU}}{\beta_V},\qquad
\eta=\frac{\eta_d}{|\beta_V|},\qquad
\lambda=\frac{\lambda_d}{|\beta_V|}.
\label{eq:normalized}
\end{equation}
This heteroskedastic Gaussian likelihood preserves the four components in Equation~\ref{eq:choice-probit}: its mean separates value sensitivity from a relative-uncertainty shift, and its variance separates uncertainty-independent choice noise from dispersion that grows with total uncertainty. The fitted \(\eta\) captures the Thompson-like signature across trials; without repeated identical prompts and independent thinking draws, it does not by itself establish within-stimulus Thompson sampling.

We fit the hybrid model separately for each model, mode, and generation condition by maximum likelihood. We also compared it with nested value-only, UCB-like-only, and Thompson-like-only specifications using \(\mathrm{BIC}=k\log n-2\log\mathcal{L}\). All reported \(\gamma\), \(\eta\), and \(\lambda\) estimates come from the hybrid specification so that each quantity has the same interpretation across conditions.

Normalization by \(\beta_V\) puts the three effects in value-equivalent units. It also exposes a genuine identification boundary: if \(\beta_V=0\), the normalized parameters cannot be defined. We therefore omit the affected GPT-OSS 120B non-thinking fit from paired normalized summaries rather than replacing it with a sentinel value. Other analyses that do not divide by \(\beta_V\) retain all ten models.

\subsection{Behavioral Regressions}

For each outcome \(y_i\), the reported coefficient analysis used
\begin{equation}
y_i=\alpha+\beta_V|V_i|+\beta_{RU}|RU_i|
  +\beta_{TU}(TU_i-TU_{\mathrm{ref}})+\epsilon_i.
\label{eq:behavior-regression}
\end{equation}
The outcomes were the magnitude of the action-logit difference \(\lvert d_i\rvert\), reported confidence \(C_i\), and log-transformed thinking length \(\log(n_{\mathrm{thinking}}/100)\). The thinking-length model used a right-censored Gaussian likelihood at the maximum length; non-thinking trials have no thinking-length outcome. The centering constant \(TU_{\mathrm{ref}}\) changes the intercept but not the uncertainty coefficient.

\subsection{Reported-Confidence Correlations}

For this analysis, \(a_i\) denotes the legal action sampled at temperature \(1\) from the two action scores and used to condition the reported confidence response. Define \(z_i=1\) when \(a_i=L\) and \(z_i=-1\) when \(a_i=R\). We characterized two marginal associations using
\begin{equation}
\begin{aligned}
E_i&=\frac{z_iV_i}{TU_i},&
M_i&=z_i d_i,
\end{aligned}
\label{eq:chosen-evidence}
\end{equation}
where \(E_i\) is evidence favoring the sampled action and \(M_i\) is its action-logit margin. We computed the Spearman correlation of reported confidence \(C_i\) with each quantity within every model and mode, then summarized those correlations across models. Because these are separate marginal correlations rather than a joint model or formal contrast, they do not establish that the mode difference is specific to task evidence or independent of the action-logit margin. The primary action analyses continued to use the unsampled signed action-logit difference \(d_i\).

\subsection{Aggregate Uncertainty}

Unless a caption states otherwise, every point or bar in the reported result figures is an unweighted mean across the included models, and every error bar is the sample standard deviation across models divided by the square root of the number of models. The cognitive mode comparison uses nine paired models because one non-thinking fit yields unidentified normalized parameters. Behavioral regressions, reported confidence analyses, reported-confidence correlations, thinking-length summaries, and thinking-generation sweeps use ten models. These error bars describe between-model variation and should not be interpreted as posterior intervals or within-model confidence intervals.

\section{Individual-Model Results}

\begin{figure*}[t]
\centering
\includegraphics[width=0.94\textwidth]{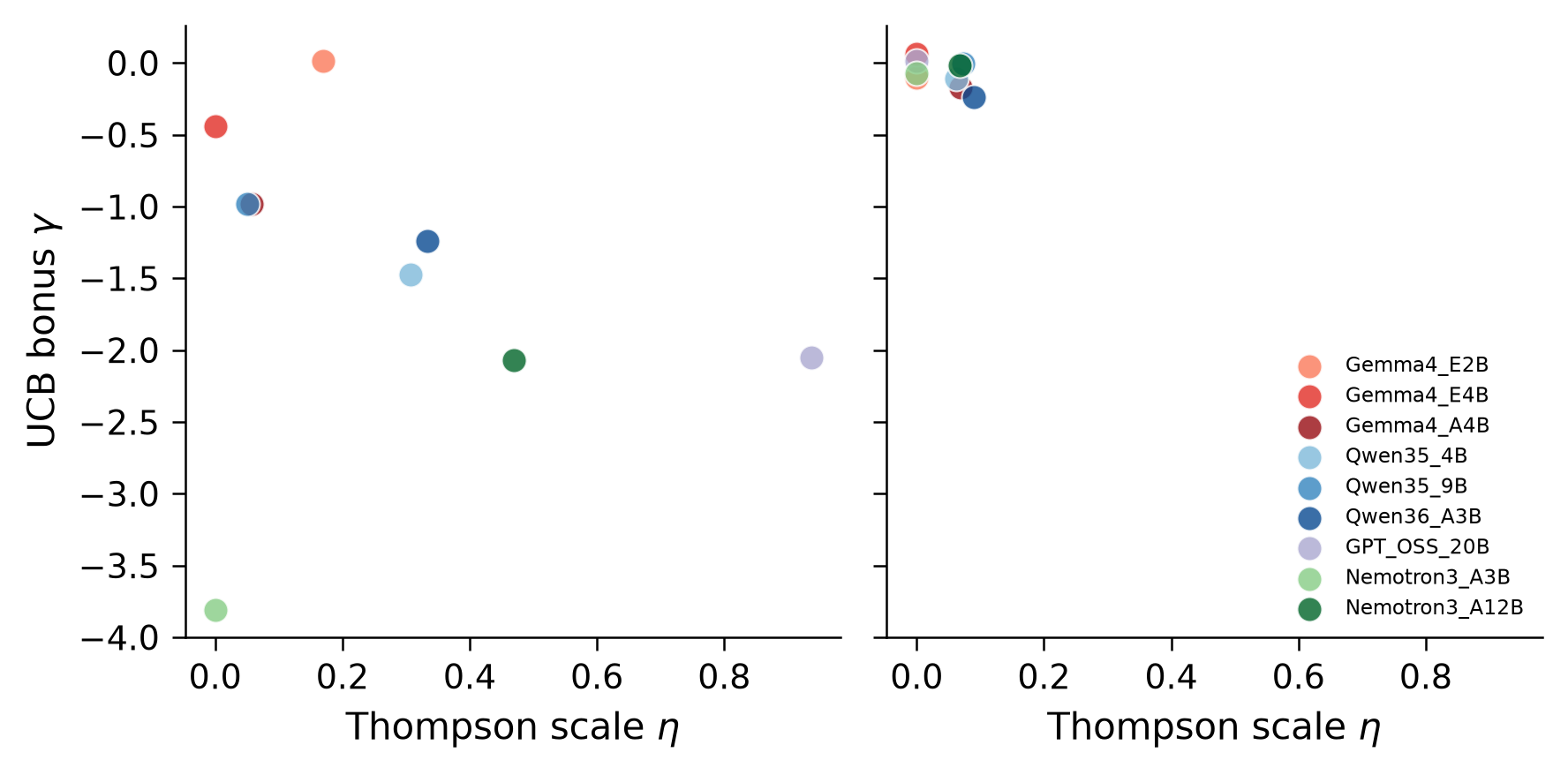}
\caption{Individual-model exploration estimates in non-thinking (left) and default-thinking (right) modes. Each point shows one model's normalized continuous-logit hybrid-model estimates of Thompson-like exploration \(\eta\) (choice variability that increases with total uncertainty) and UCB-like exploration \(\gamma\) (a preference for the less-known arm), fitted across 600 Horizon trials; colors identify the models in Table~\ref{tab:models}, and the panels share axes. The figure includes the nine models with identifiable reward-unit parameters in both modes. GPT-OSS 120B is omitted because its fitted non-thinking \(\beta_V\) was zero, leaving its normalized parameters unidentified. Points are model-specific estimates; no error bars are shown.}
\label{fig:supp-individual-eta-gamma}
\end{figure*}

Across the nine models with identifiable reward-unit parameters in both modes, non-thinking estimates were widely dispersed, whereas default-thinking estimates clustered near zero for both \(\eta\) and \(\gamma\) (Fig.~\ref{fig:supp-individual-eta-gamma}). This model-level pattern supports the aggregate conclusion that thinking did not add coherent UCB-like or Thompson-like exploration.

\begin{figure*}[t]
\centering
\includegraphics[width=0.94\textwidth]{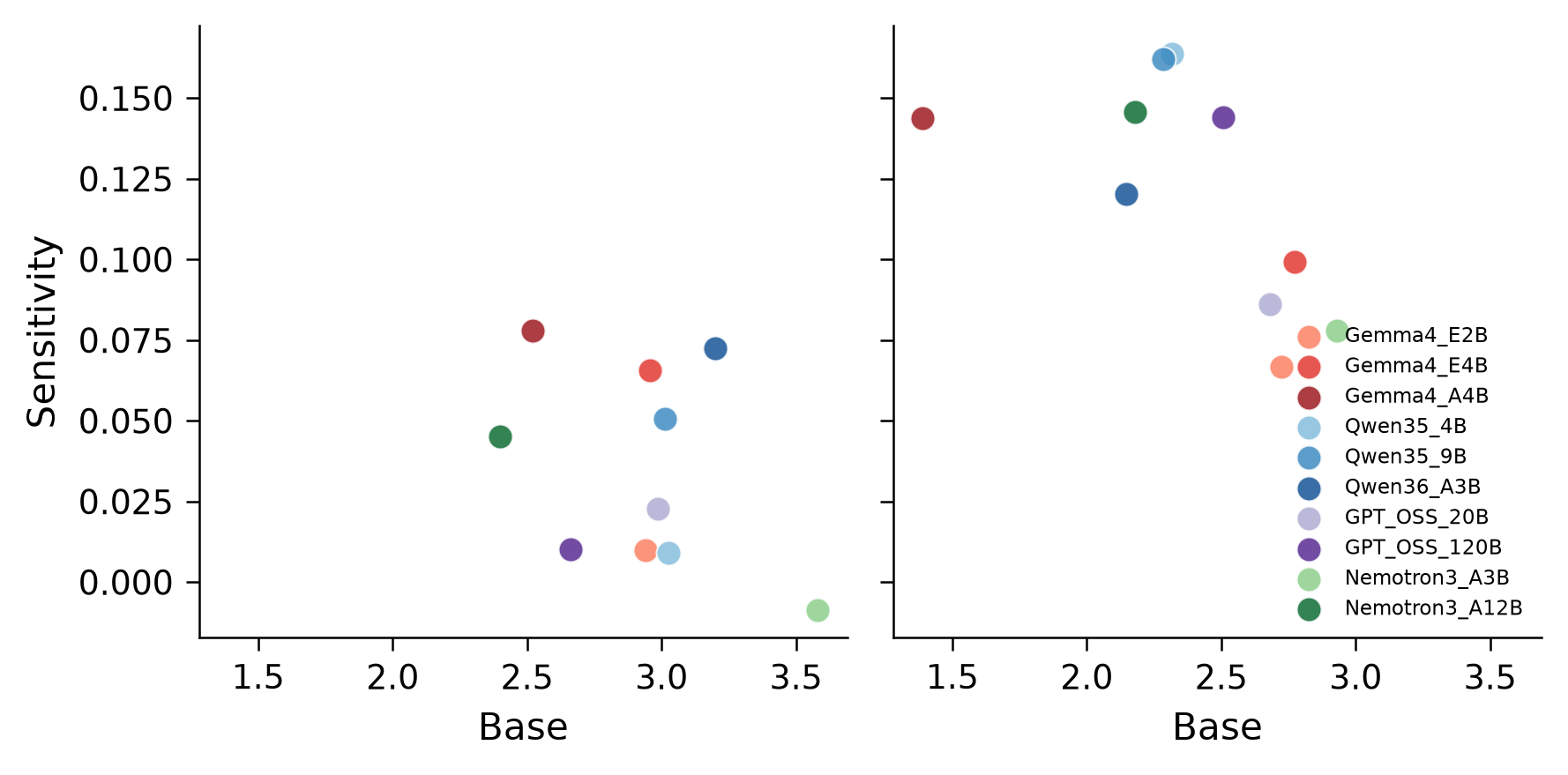}
\caption{Individual-model reported confidence fits in non-thinking (left) and default-thinking (right) modes. For each model and mode, reported confidence on the common 1--4 scale was fit across 600 Horizon trials as \(C_i=b+s\lvert V_i\rvert\); point coordinates give the intercept (Base) and slope (Sensitivity). Colors identify the ten models in Table~\ref{tab:models}, and the panels share axes. Thinking lowered the fitted intercept and increased value sensitivity for every model. Points are model-specific estimates; no error bars are shown.}
\label{fig:supp-individual-confidence-fit}
\end{figure*}

The reported confidence pattern was more consistent across models. All ten models had a lower fitted reported confidence baseline and greater sensitivity to absolute value difference under default thinking, although the size of each shift varied (Fig.~\ref{fig:supp-individual-confidence-fit}).

